# Qualitative Probabilistic Networks for Planning Under Uncertainty

Michael P. Wellman[*]
MIT Laboratory for Computer Science

## Abstract

Bayesian networks provide a probabilistic semantics for qualitative assertions about likelihood. A qualitative reasoner based on an algebra over these assertions can derive further conclusions about the influence of actions. While the conclusions are much weaker than those computed from complete probability distributions, they are still valuable for suggesting potential actions, eliminating obviously inferior plans, identifying important tradeoffs, and explaining probabilistic models.

## 1 Introduction

In the traditional planning paradigm, the effects of actions are represented as logical functions of the state in which they are applied. In practice, the representation is restricted by difficulties of expression and deductive limits of the reasoner. When the planner's knowledge of the environment is uncertain, the representation problem is substantially more complicated. Because the result of an action is not uniquely determined by the model, it is necessary to enumerate the set of possible results that can obtain along with their respective likelihoods.

One class of representations is the *probabilistic model*. Probabilistic models have the virtue of completeness; the output is a full distribution of outcomes for any plan of interest. Furthermore, the model has a decision-theoretic semantics that can be used to validate and interpret its component parts. However, opponents of this approach argue that the completeness and semantic adequacy are an illusion, since structurally correct and numerically precise models cannot be constructed in practice. New uncertainty calculi and various deterministic approaches to uncertainty have been proposed as alternate methodologies.

The approach described here attempts to answer some of the concerns of non-probabilists without abandoning probabilistic semantics. Essential characteristics of the influence of actions on the world can often be captured with qualitative assertions that are much easier to specify than complete probabilistic models. Conclusions derived from qualitative abstractions can be justified by decision theory, with the added assurance that they are not an artifact of unreasonable precision in our specification of the model. Of course, such conclusions will be weaker than those drawn from completely specified models, and are not sufficient in general to select a uniquely optimal plan. Nevertheless, qualitative models will often reduce the space of admissible plans, help to focus the search for good plans, identify central tradeoffs in the decision problem, and improve explainability.

## 2 Probabilistic Networks

Two related graph-based formalisms that have been advocated for computer representation of probabilistic knowledge are Pearl's *Bayesian networks* [9] and the *influence diagrams* of Howard and Matheson [6]. Graph representations are computationally attractive and have conceptual advantages in their focus on dependencies among the probabilistic variables [10].

Bayesian networks and influence diagrams both encode probabilistic models as directed graphs,

---

[*]This research was supported by National Institutes of Health Grant No. R01 LM04493 from the National Library of Medicine and National Institutes of Health Grant No. R24 RR01320 from the Division of Research Resources.

with the nodes representing uncertain variables and the links denoting probabilistic dependence. Within each node is a table recording the distribution of the node's values given each combination of values for its direct predecessor nodes. Distributions of nodes of interest under various scenarios or decisions may be computed through propagation or graph reduction techniques. Influence diagrams also include specially-designated decision nodes and informational links to indicate which chance variables are known at the time decisions are made. A single value node indicates the utility of outcomes represented in the network.

Because decision-making is the focus of this work, I will discuss the methodology within the influence diagram paradigm. The presentation of manipulation techniques below closely parallels Shachter's description of the algorithm for evaluating numeric influence diagrams [12].

## 3 Qualitative Influences

### 3.1 Definitions

Qualitative probabilistic networks replace the conditional distribution table within each node with a specification of the *direction* of a predecessor's influence. Consider an influence diagram with binary chance nodes.[1] Lowercase letters are variables corresponding to each node, and uppercase letters are literals denoting the proposition's truth or falsity (e.g., $A$ and $\bar{A}$). Often the literal $X$ serves as an abbreviation for the proposition $x = X$. We say that node $a$ positively influences node $b$— denoted $I^+(a,b)$—if and only if

$$\forall x \ \Pr(B|Ax) \geq \Pr(B|\bar{A}x) \quad (1)$$

Equivalently,

$$\forall x \ \Pr(\bar{B}|Ax) \leq \Pr(\bar{B}|\bar{A}x) \quad (2)$$

Here, $x$ ranges over the propositional formulas consistent with both $A$ and $\bar{A}$. Negative and zero influences are defined analogously. An unknown dependence between $a$ and $b$ is written $I^?(a,b)$.

[1]The extension to multi-valued chance variables is straightforward.

It is also useful to provide a notation for conditional influence assertions. We can assert that $a$ positively influences $b$ given $y$:

$$I^+(a,b,y) \equiv \forall x \ \Pr(B|Ayx) \geq \Pr(B|\bar{A}yx) \quad (3)$$

Such an assertion says nothing about $a$'s influence on $b$ when $y$ does not hold. Notice that unconditional influence is just a special case of conditional influence where $y = \mathbf{true}$.

These influences can be depicted graphically by placing the appropriate direction notation on the links between nodes. Figure 1 displays a fragment of a qualitative probabilistic network consisting of one node influenced by several others. Conditional influences are indicated by writing the condition after the direction, separated by a vertical bar. A link may contain several influences, but the conditions for each must be mutually exclusive.

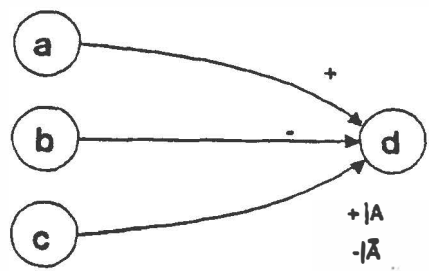

Figure 1: Graphical notation for qualitative influences.

Assertion of qualitative influences is far weaker than the conditional distribution table specified for complete influence diagrams. In general, the influences acting on a node induce a partial order on the conditional probabilities for the node's event given its predecessors. For example, in the diagram of Figure 1, the assertions imply $\Pr(D|ABC) \geq \Pr(D|\bar{A}BC)$. The partial order determined by the influences in this case is shown in Figure 2.

This partial order could be further constrained by assertions of pairwise influence. For example, we might assert that for any $x$, $\Pr(D|ABx) \geq \Pr(D|\bar{A}\bar{B}x)$. Joint influence assertions are not considered here, but the techniques can be extended to accommodate them.

Influences on the value node of an influence diagram are defined in a similar manner. The assertion $I^+(a,u,y)$ means that node $a$ has a positive

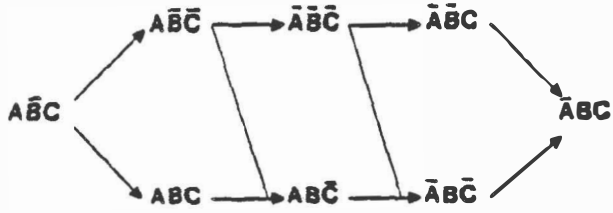

Figure 2: Partial order on the conditional probabilities for $D$ given $a$, $b$, and $c$. An arrow from $x$ to $y$ indicates that $\Pr(D|x) \geq \Pr(D|y)$.

influence on utility $u$ given $y$.

$$I^+(a, u, y) \equiv \forall x\ u(x, A, y) \geq u(x, \bar{A}, y) \quad (4)$$

where $x$ is an assignment of values to nodes other than $a$ and those in $y$. $I^+(u, a, y)$ is undefined.

### 3.2 Graph Manipulations

Like numeric influence diagrams, qualitative influence diagrams are "evaluated" by successive removal of nodes from the network. Any chance node with a single direct successor may be removed by "splicing" its direct predecessors to its successor, determining appropriate influences for the new links. These new influences can be computed from the influences on the old links according to a simple qualitative algebra. The operations for combining influences are similar to those for combining qualitative measures in other applications of qualitative reasoning in AI (e.g. see Kuipers [7]).

### Influence Chains

Consider the simple chain shown in Figure 3. If there are no other influences in the network on $b$ or $c$, then it is clear that $I^+(a,b) \wedge I^+(b,c) \Rightarrow I^+(a,c)$.

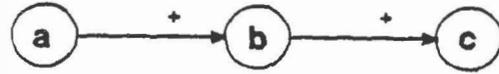

Figure 3: A simple influence chain.

To verify that this also holds if there are other influences on $b$ and $c$ (possibly including additional indirect paths from $a$ to $c$), see Figure 4.

It is convenient to define an operator, $\otimes$, for combining chains of qualitative influences. The complete definition for $\otimes$ is provided in Table 1. It is apparent from the table that $\otimes$ is just sign

---

Let $x_0$ be any assignment of values to the other nodes in the network. We have

$$\begin{aligned}\Pr(C|Ax_0) &= \Pr(BC|Ax_0) + \Pr(\bar{B}C|Ax_0) \\ &= \Pr(C|ABx_0)\Pr(B|Ax_0) + \Pr(C|A\bar{B}x_0)\Pr(\bar{B}|Ax_0)\end{aligned} \quad (5)$$

Because $a$ and $c$ are conditionally independent given $b$ and $x$,

$$\Pr(C|Ax_0) = \Pr(C|Bx_0)\Pr(B|Ax_0) + \Pr(C|\bar{B}x_0)\Pr(\bar{B}|Ax_0) \quad (6)$$

and using $\Pr(B|Ax) + \Pr(\bar{B}|Ax) = 1$,

$$\Pr(C|Ax_0) = \Pr(B|Ax_0)\left[\Pr(C|Bx_0) - \Pr(C|\bar{B}x_0)\right] + \Pr(C|\bar{B}x_0) \quad (7)$$

Similarly,

$$\Pr(C|\bar{A}x_0) = \Pr(B|\bar{A}x_0)\left[\Pr(C|Bx_0) - \Pr(C|\bar{B}x_0)\right] + \Pr(C|\bar{B}x_0) \quad (8)$$

Given the inequalities expressed by the influence assertions,

$$\forall x\ \Pr(C|Bx) \geq \Pr(C|\bar{B}x) \text{ and } \forall x\ \Pr(B|Ax) \geq \Pr(B|\bar{A}x) \quad (9)$$

it follows that $\Pr(C|Ax_0) \geq \Pr(C|\bar{A}x_0)$. Since $x_0$ was chosen arbitrarily, we have $I^+(a, c)$. ∎

Figure 4: A demonstration that $I^+ \otimes I^+ = I^+$.

---

multiplication with conjunction of the conditioning propositions. Combination of influences is thus associative and commutative.

| $\otimes$ | $+$ | $-$ | $0$ | $?$ |
|---|---|---|---|---|
| $+$ | $+$ | $-$ | $0$ | $?$ |
| $-$ | $-$ | $+$ | $0$ | $?$ |
| $0$ | $0$ | $0$ | $0$ | $0$ |
| $?$ | $?$ | $?$ | $0$ | $?$ |

Table 1: The $\otimes$ operator for combining influence chains. For example, $I^{+|x} \otimes I^{-|y} = I^{-|x \wedge y}$.

### Influence Addition

The derivation above assumed that there were no direct links between $a$ and $c$.[2] But even if there are such links, it still makes sense to say that $a$ positively influences $c$ *along this particular path*. After removing node $b$, however, we are left with two direct links from $a$ to $c$. This situation is illustrated in Figure 5.

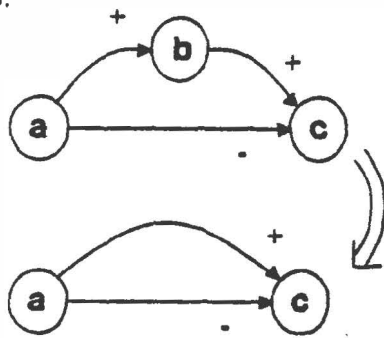

Figure 5: Removing node $b$ leaves two paths from $a$ to $c$.

Parallel influences can be combined in a way analogous to the serial influence chains described above. The influence addition operator, $\oplus$, is defined by Table 2. Note that the situation of Figure 5 is ambiguous; we cannot determine the direction of the overall influence of $a$ on $c$.

---

[2]If there were, then the conditional independence assumption necessary for going from equation (5) to equation (6) would not be valid.

| $\oplus$ | $+$ | $-$ | $0$ | $?$ |
|---|---|---|---|---|
| $+$ | $+$ | $?$ | $+$ | $?$ |
| $-$ | $?$ | $-$ | $-$ | $?$ |
| $0$ | $+$ | $-$ | $0$ | $?$ |
| $?$ | $?$ | $?$ | $?$ | $?$ |

Table 2: The $\oplus$ operator for combining parallel influences.

### Arc Reversal

Shachter [12] shows that during influence diagram evaluation there is always a candidate node for removal, sometimes requiring arc reversals first. It is possible to reverse arcs in a qualitative network as well, with similar requirements for updating the predecessor relations for the nodes involved. The reversed influence has the same direction as the original one, and each node adds the direct predecessors of the other to its own. New predecessor influences are computed from the old values.

### Decision Nodes

Decision nodes may be removed from the network if the optimal choice is apparent from the influences in the reduced diagram. For example, if the decision node is an unambiguous positive or negative influence on utility, then the choice is obvious. Given particular values for the informational predecessors, the decision can be determined if the conditional influence of the decision on utility given those values is unambiguous.

In general, decision nodes cannot be easily removed. Such cases require more powerful analysis techniques, like those illustrated below.

## 4 An Example: The Generic Test/Treat Decision

Consider the following rather vague description of a medical decision problem:

> Patient $P$ might have the undesirable disease $D$. Patients with $D$ are sometimes "cured," which is preferred over uncured disease. A treatment exists which improves the likelihood that $P$ will be cured, but this treatment may result in unpleasant side-effects. There is also a test for which a positive result is somewhat indicative of disease $D$. The test carries some risk of undesirable complications.

The information presented thus far is obviously insufficient for deciding the fate of poor patient $P$. Nevertheless, it conveys enough of the structure of the decision problem to construct the qualitative influence diagram depicted in Figure 6.

As the diagram indicates, utility (the hexagon node) is a function of the four variables $c$, $d$, $y$, and $z$. The result is conditionally independent of the decisions (square nodes) and the test result given those four values. Note that cure only influences utility in the presence of disease.

The influence of the test result on the likelihood of disease is similarly conditioned on whether or not the test was performed. Because of this conditioning, node $d$ depends probabilistically on node $t$. But since this dependence is totally described by the condition on the $r$ influence, we do not need a separate influence for $t$. The dependence is indicated on the diagram by a dashed line to avoid portraying a redundant influence in the network.

The first step in evaluating the diagram is to apply the graph manipulations described in section 3.2 to reduce the model as much as possible. In our example, removing nodes $c$, $z$, and $y$ and reversing the influence between $r$ and $d$ leaves us with the diagram of Figure 7.

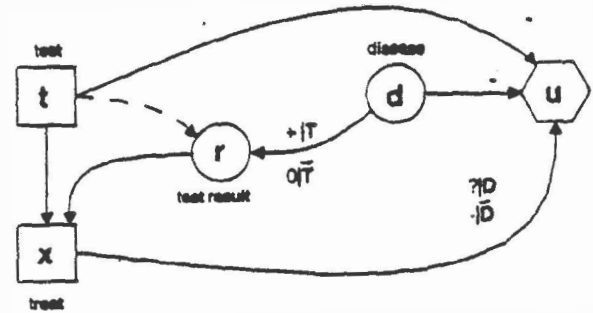

Figure 7: The reduced influence diagram.

## 4.1 Planning

Before proceeding further, let us consider our objectives in this decision problem. We are certainly not going to derive a unique decision from this information. Instead, an analysis of the model serves two main purposes: to separate the sensible plans from the senseless ones and to identify the indeterminacies of the model which are most important for resolving the decision at hand. The remainder of this discussion is devoted to the first purpose.

A naive planning program or influence diagram evaluator considers all syntactically valid plans. In a representation that consists of a collection of available actions, the set of syntactically valid plans is simply all sequences of actions. If there are uncertain variables that may become known along the way, then plans must include contingencies.

In the reduced diagram of Figure 7, the set of syntactically valid plans is defined by the combinations of $t$ and $x$, with $x$ expressed as policies given $t$ and $r$. This makes for a total of eight dis-tinct strategies. While many of these combinations are reasonable, others (more than half) either violate common sense or are not coherent plans at all. For example, the combination of $T$ with the treatment policy $x(r) = X$ seems like a ridiculous strategy, because we are performing the test but then treating the patient regardless of the result. The strategy of not doing the test but treating only if positive is also syntactically valid in this model, though it has no coherent meaning in the domain.

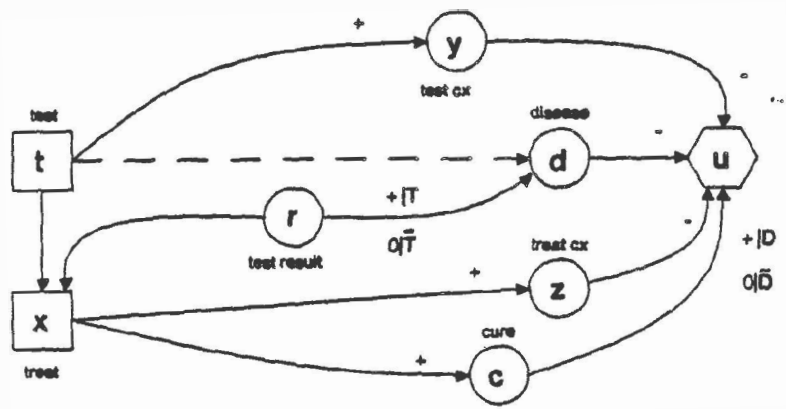

Figure 6: Qualitative influence diagram for the generic test/treat decision.

In the following section we will see that—at least in this example—the intuitively nonsensical plans correspond exactly to those not admissible under qualitative influences.

### 4.2 Determining Admissibility

A variety of techniques are available for computing admissibility within a qualitative probabilistic network. This section explores a few of these by examining their conclusions about the generic test/treat example.

#### Hypothetical Optimality

The "hypothetical optimality" technique explores the space of possible plans by postulating that the optimal strategy includes particular components. For example, we might start on the current problem by assuming that the best plan includes a test operation ($t = T$). Examining the diagram of Figure 7, we see that node $t$ has two paths to the value node: an unambiguously negative path due to the effect of test complications, and an indeterminate path which includes the informational link from the test result to the treatment decision.

If testing is indeed optimal, then the indeterminate path must actually have a positive influence on utility. Otherwise, performing the test would have an overall negative value. An informational path can only have positive value if a downstream decision depends on the value of a chance node. That is, the value of $x$ which is preferred given $R$ is not the preferred value given $\bar{R}$. Thus, one of the following two conditions must hold:

$$I^+(x, u, TR) \wedge I^-(x, u, T\bar{R}) \qquad (10)$$

$$I^-(x, u, TR) \wedge I^+(x, u, T\bar{R}) \qquad (11)$$

The second conjunct of condition (11) implies that negatives should be treated. But since $\Pr(D|R) \geq \Pr(D|\bar{R})$, this clause contradicts the first. Thus, given that testing is optimal, the strategy of treating only the positives is best.

Similar hypothetical reasoning can show that if treating negatives is optimal then it does not make sense to perform the test. Proofs of this type rule out all strategies in this example except

- no test, no treatment
- test, treat if positive
- no test, treat (empiric therapy)

#### Stochastic Dominance

Though the previous technique has substantial intuitive appeal, it is not clear that such nice admissibility results can always be derived via simple inspection. Therefore, we seek a more systematic approach that will eliminate as many nonsensical strategies as possible.

Figure 8 displays the partial order on utility induced by the influences in the reduced influence diagram. We can use this partial ordering to prove strategies inadmissible through a generalized form of *first-order stochastic dominance* [15]. Strategy $A$ dominates strategy $B$ in the first order if for every possible outcome, the likelihood of getting that outcome or a worse one is greater for $B$ than for $A$. This can be generalized to partially ordered outcomes by requiring that the condition hold within directed paths in the order graph, being careful to avoid double-counting of outcomes.

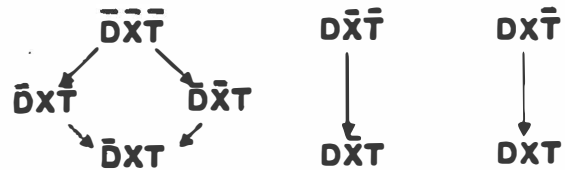

Figure 8: Partial order on outcome desirability determined from the diagram.

We can rule out a particular strategy by finding another feasible strategy that dominates it according to this criterion. To perform this comparison, we determine the outcomes of each strategy given the various possibilities for relevant chance nodes in the network. For example, Table 3 is a case analysis of the two syntactically valid strategies "no test, no treat" and "test, no treat."

| Strategy | Case | Prob | Outcome |
|---|---|---|---|
| $t = \bar{T}, \ x = \bar{X}$ | $D$ | $\Pr(D)$ | $D\bar{X}\bar{T}$ |
|  | $\bar{D}$ | $\Pr(\bar{D})$ | $\bar{D}\bar{X}\bar{T}$ |
| $t = T, \ x = \bar{X}$ | $D$ | $\Pr(D)$ | $D\bar{X}T$ |
|  | $\bar{D}$ | $\Pr(\bar{D})$ | $\bar{D}\bar{X}T$ |

Table 3: Case analysis of two strategies.

The table records the outcome for each case, along with its probability (symbolic). Notice that the value of node $r$ is irrelevant here, since there is no path to utility except through an informational link which is not being used. Looking at Table 3, it is easy to see that the first strategy dominates the second. For each case, the first strategy has an outcome which is preferred (based on the partial order) to the corresponding outcome in the second strategy, which occurs with equal probability. This ranking does not depend at all on the prevalence of the disease.

By an almost identical argument, the strategy "test, treat" is dominated by the empiric therapy strategy.

Unfortunately, pairwise dominance testing is not sufficient to rule out all inadmissible strategies. In our example, the strategy "test, treat iff negative" is not dominated by any of the other strategies under consideration. While the strategy of treating only the positives may seem to make more sense, it is actually the inferior choice when, for example, the test is very insensitive and the treatment complications are relatively unlikely or benign.

To prove that treating the negatives is a suboptimal policy, it is necessary to show that there is always some strategy that is preferred, though the superior strategy may vary from case to case. In the scenario mentioned above where this policy is better than treating the positives, it is clear that there are other strategies that would be even better. Indeed, this is always the case in our example. For this kind of situation, dominance proofs take the form

$$S_1 \succeq S_2 \Longrightarrow S_3 \succeq S_1 \qquad (12)$$

When (12) holds, we say that $S_1$ is dominated with respect to $S_2$ and $S_3$.

Three-way (or $k$-way) dominance proofs are more cumbersome to construct, however. A technique that might be computationally more direct is to demonstrate suboptimality by comparison with *mixed strategies* [4]. A mixed strategy is simply a probabilistic combination of feasible deterministic strategies. For our example, consider

**Strategy 9:** No test, treat if heads on an $\alpha$ coin flip.

It can be shown that strategy 9 with $\alpha = \Pr(\bar{R}|D)$ is guaranteed to dominate the treat on negative strategy, even though no nonrandom strategy is always superior. Since mixed strategies are always suboptimal [11], our problematic strategy of treating only the negatives may be ruled inadmissible.

## 5  Conclusions

Within a probabilistic network formalism, we can provide a precise probabilistic semantics for qualitative influences among uncertain variables. A qualitative algebra for combining influences in the network can be used within an algorithm for isolating the influences of interest, such as the overall influence of decisions on expected utility.

Given a reduced influence diagram, several techniques are available for determining the admissible strategies among the syntactically valid plans. The source of indeterminacy in a failed dominance proof may be a reliable indicator of the importance of further information and assumptions to a problem solver capable of generating more detailed models.

The scheme presented here is intended to form only part of a comprehensive planning program. Although these kinds of qualitative influences are not sufficient to resolve true tradeoffs, the techniques described can be useful in planning under uncertainty for a variety of reasons.

First, the qualitative reasoner can work in tandem with a more traditional planner or decision algorithm, noticing the "obvious" conclusions before precise information is supplied. This argument suggests that a program that rules out "test but do not treat" based on a lower numeric expected utility is missing the point, because the conclusion does not depend on the absolute values of any quantities in the model. Recognition of the minimal assumptions necessary for a result provides for more coherent and compelling explanations than those generated under complete information.

Second, the mechanisms described here may be "scaled up" to handle more precise information when it is available, with decision-making power varying smoothly within the continuum of model specification. Expansion of indeterminacy with increased detail can be avoided somewhat by maintaining the more abstract influences while exploring the more detailed structure. Thus, this work is complementary with other recent AI work on reasoning with incomplete probabilistic information [2,5,8] as well as incomplete utility specification [14].

Finally, the formalism provides a means for exploring probabilistic interpretations of other schemes that use qualitative influence among uncertain variables. Work on "causal networks" (e.g. CASNET [13]) and Cohen's endorsement approach [1] fall in this category.

Further development and integration will buttress these arguments. Current work includes application of these techniques to analyze and critique models generated by human decision analysts [3].